\documentstyle[11pt,epsf]{article}
\newcommand{\be}{\begin{equation}}
\newcommand{\ee}{\end{equation}} 
\newcommand{\eei}{\end{equation}\indent\indent}
\newcommand{\bc}{\begin{center}}
\newcommand{\ec}{\end{center}}
\newcommand{\ber}{\begin{eqnarray}}
\newcommand{\ear}{\end{eqnarray}}
\newcommand{\ba}{\begin{array}}
\newcommand{\ea}{\end{array}}
\newcommand{\vb}{\verb}

\def\case#1/#2{\textstyle\frac{#1}{#2} }
\begin{document}
\title{Name Strategy:  Its Existance and Implications}
\author{Mark D. Roberts, \\\\
Department of Mathematics and Applied Mathematics,\\
University of Cape Town,\\
Rondbosch 7701,\\
South Africa\\\\
roberts@gmunu.mth.uct.ac.za} 
\date{\today}
\maketitle
\vspace{1.0truein}
\bc Mathematical Reviews Subject Classification: \ec
\bc  92J10,  04A30,  68T30,  03B46. \ec
\bc Association for Computing Machinery Classification:  \ec
\bc  I.2.6;  J.4;  I.2.7  \ec 
\bc 32 pages,  2 ascii diagrams. \ec
\newpage
\begin{abstract}
It is argued that colour name strategy,  object name strategy,
and chunking strategy in memory are all aspects of the same general 
phenomena,  called stereotyping,  and this in turn is an example of a 
know-how representation.   Such representations are argued to have their 
origin in a principle called the {\it minimum duplication of resources}.   For 
most the subsequent discussions existence of colour name strategy suffices.
It is pointed out that the Berlin-Kay universal partial ordering of colours 
and the frequency of traffic accidents classified by colour are surprisingly 
similar;  a detailed analysis is not carried out as the specific colours 
recorded are not identical.   Some consequences of the existence of a name 
strategy for the philosophy of language and mathematics are discussed:
specifically it is argued that in accounts of truth and meaning it is 
necessary throughout to use real numbers as opposed to bivalent quantities;
and also that the concomitant label associated with sentences should not be 
of unconditional truth,  but rather several real valued quantities associated
with visual communication.   The implication of real valued truth quantities
is that the {\bf Continuum Hypothesis} of pure mathematics is side-stepped,  
because real valued quantities occur {\it ab initio}.   The existence of name 
strategy shows that thought/sememes and talk/phonemes can be separate,  
and this vindicates the assumption of thought occurring before talk used
in psycholinguistic speech production models.
\end{abstract}
\newpage
\section{Contents}
\bc {\bf Introduction} \ec
2.1 Motivation   
2.2 Diagram:  Name Strategy in the Representation Hierarchy 
2.3 Sectional Contents   
2.4 Stereotypes compared to Other Representations.        
2.5 Minimal Duplication of Resources.\\ 
{\bf Part One Existence}
\bc{\bf Exitence of Name Strategy in Colour Perception}  \ec
3.1 Berlin-Kay Colour Ordering          
3.2 The Sapir-Worf Hypothesis          
3.3 Name Strategy              
3.4 Remarks on Colour Perception    
\bc {\bf Existence of Name Strategy in Object Perception} \ec    
4.1 Object Perception                       
\bc {\bf Memory Chunking} \ec
5.1 Recalling Recoded Events   
5.2 Memory for Chinese Words and Idioms.\\
{\bf Part Two Implications}
\bc{\bf Implications for Traffic Accidents}\ec 
6.1 Colour Perception in Traffic Accidents
\bc {\bf Implications of Name Strategy for Psycholinguistics}   \ec
7.1 The Interaction Model        
7.2 Serial and Connectionist Models   
7.3 Language Production     
7.4 Sememes and Phonemes verses Thought and Talk   
7.5 Justification of Prior Thought    
\bc {\bf Implications for the Philosophy of Language} \ec
8.1 Radical Interpretation        
8.2 Accuracy of Correspondence 
8.3 Quantity of Transferred Information when in Re-coded Form 
8.4 Circuitous Correspondence  
8.5 Flagging or the Extralinguistic Assignment Problem    
8.6  Flagging by Countenance    
\bc {\bf Implications for the Philosophy of Mathematics} \ec
9.1 The Continuum Hypothesis and the Segmentation Problem 
9.2 Probabilistic and Fuzzy Tarski Truth Theory          
\bc {\bf Summary} \ec
10.1 Peroration        
10.2 Conclusion 
\bc {\bf Acknowledgements} 
~~~~~~~~~~~~~{\bf References} \ec
\section{Introduction}
\subsection{Motivation}
\label{sec:moti}
Name strategy is a type of deployment of a representation whose 
relationship to other and general representations is shown by 
the hierarchial diagram \ref{sec:dia} below.   
Name strategy has particularly well-defined features clarifying the 
related properties of other entries in the diagram.  Roughly the entries in 
the diagram can be explained as follows.   {\it Representation} 
has no all-encompassing formal definition.   
In artificial intelligence it roughly means a set
of conventions about how to describe things;  it is a loose term to be kept in 
mind when creating programs,  languages etc.,  which entails keeping track of 
how knowledge or data or anything else is codified and whether this involves 
some condensation of the original material.   It splits into two parts:
i){\it know-that representations} 
- usually these hold factual knowledge or data, 
and typically are the subject of information theory,   
ii) and {\it know-how representations}
- usually these hold inference knowledge in the form of 
programs and languages,  and typically are the subject of instruction theory.
For animals,  as opposed to machines,  {\it representation} again means the 
codified form of knowledge and data;  again it splits into two parts: 
{\it know-that representations} which are stores of encyclopedic knowledge,  
and {\it know-how representations} which are essentially skills that can 
be deployed.   {\it Stereotyping} loosely means collating information 
into a labeled set or template which can be readily (re-)used;  
properly this is cognitive stereotyping as opposed to social 
stereotyping,  Stewart et al (1979) \cite{bi:stewart}.   Examples of this are 
actions such as walking;  an example which allows quantitative 
measurements is {\it memory chunking},  the existence of which 
has been readily demonstrated by many experiments.   {\it Name strategy} means 
associating a given word with a stereotype.   This splits into two types: 
i) where the word activates a switch,  the switch is perhaps innate,  
another example of a switch is pro-drop in linguistics,  
ii)where there is no switch.   The paradigmatic,  and best studied,  
example of the first of these is colour name strategy;  in which the 
focal colours are progressively labeled by names as a culture advances 
until all eleven are named.   Object name strategy is an example of the 
second.   The purpose of the present paper is to describe the empirical 
justification for these representations and their implications for 
psychology,  philosophy,  and mathematics.
\newpage
\subsection{Diagram}
\label{sec:dia}
\begin{verbatim}

                                                                         
2.2  Diagram:   Name Strategy in the Representation Hierarchy


                        Representations
                        /            \
                       /              \
                      /                \
                    \/                  \/
           Know-how Representations      Know-that Representations
                                 \
                                  \
                                   \
                                    \/
                                   Stereotyping
                                   /          \
                                  /            \
                                 /              \       
                               \/                \
                    Name Strategy                 \
                   /             \                 \
                  /               \                 \
                 /                 \                 \
               \/                   \/                \/
Colour Name Strategy    Object Name Strategy        Memory Chunking
                                                                           

\end{verbatim}
\newpage
\subsection{Sectional Contents}
\label{sec:sc}
Section I establishes a framework which allows discussion of the nature
of representations;  it is advocated that there is minimal duplication of 
these.  
The existence of colour name strategy is discussed 
in section \ref{sec:colour}.
The similarity of the idea of adopting a name strategy and creating a 
representation discussed in section \ref{sec:stereo}.  
Colour name strategy is taken to be the paradigmatic case of stereotyping 
and the relationships of object name strategy and chunking to it are 
discussed in sections \ref{sec:objpc} and \ref{sec:memch} respectively.   
Whenever colour name strategy is deployed it illustrates that thought can 
exist with out words and in principle allows a qualitative measure of this.   
In section \ref{sec:traffic} 
it is noted that the Berlin-Kay universal partial ordering for 
colours is similar to the frequency that traffic accidents occur for a car of 
a given colour.  
Section \ref{sec:radical} continues,  from Roberts (1998) \cite{bi:mdr98},  
my approach to radical interpretation.   
This theory and how I differ from the traditional 
approach is summarized in section \ref{sec:peroration}.   
Name strategy suggests at least three 
problems with radical interpretation,  they are:  i) {\it the reduced accuracy 
problem},  which is: Why have two colour names instead of one as this can 
reduce accuracy?  ii)  {\it the segmentation problem},  which is:  
How does a theory of meaning account for the real numbers, 
or why is the {\bf Continuum Hypothesis} of mathematics justified?  
iii) {\it the extralinguistic assignment problem},  
which is:  traditional radical interpretation requires extralinguistic 
information in order to assign truth values to bivalent statements;
the extralinguistic assignment problem asks:  what is the correct concomitant 
label (or flag) associated with statements and how is it induced?   
In section \ref{sec:psy} 
it is pointed out that psycholinguistic speech production models require 
{\it a-priori} structure from which speech is produced.   
The example of an interacting speech production model is investigated,  
here the starting point is a 
word in quotation marks which represents the meaning of a word.   It has long 
been argued by philosophers that a word in isolation has ambiguous meaning,  
so that there is the problem of why it is justifiable to start with the 
meaning of a word in isolation.   It is here argued that the existence of 
name strategy indicates the independence of thought and talk, and thus 
vindicates the assumption of psycholinguistic speech models,  namely that 
there is a starting point in production which is independent of words.   
Section \ref{sec:conclusion} is the conclusion.
\newpage
\subsection{Stereotypes Compared to Other Representations}
\label{sec:stereo}
The properties of colour name strategy 
are that it is a stereotype which in addition:
i) has eleven switches, see section \ref{sec:berlin},  
ii) is culturally determined,  see section \ref{sec:berlin}, 
and iii)is encapsulated.   {\it Encapsulation}  
means that once the stereotype 
has been created it cannot be adjusted and that its modes of interaction are 
few and in principle determined:  encapsulation results in a higher speed of 
processing.   Object name strategy section \ref{sec:objpc} 
and memory chuncking section \ref{sec:memch} are stereotypes which:  
i) are not switched,  ii) are determined by the environment,
iii) are encapsulated.   How does what is here called a stereotype compare to 
other representations?   {\it Semantic representations},  
Shannon (1988) \cite{bi:shannon}, are essentially codified 
and condensed information which is meant to encompass the 
``meaning'' of a portion of language.   Shannon argues that they are not a 
primary structure,  this suggests that there is no switching and meaning 
(in the form of semantic representations) is stored in a similar way to 
object perception.   Way (1991) \cite{bi:way}
allows knowledge representations to be 
general enough to include metaphor.   At this point what a representation 
is becomes entangled with what meaning is.   My view is that ``meaning'' is 
too general and nebulous a term to pin down in this way,  
aspects of ``meaning'' must be involved in colour name strategy,  
the best thing to do is classify 
aspects of meaning and see what sort of representation they entail.   Visual 
perception entails perceiving objects,  and so contains object perception as 
discussed in section \ref{sec:objpc};   Farah (1988) \cite{bi:farah}
discusses whether visual imagery engages 
some of the same representations as used in visual perception.   She finds 
neuropsychological evidence which suggests that some representations are used
for both purposes.   This dual purpose or multiuse of visual representations 
is similar to the multiuse of names in name strategy.   Irrespective of 
whether representations are innate,  the variety of use to which a given 
representation can be put suggests that there are general principles involving
the creation and operation of representations.
\newpage
\subsection{Minimal Duplication of Resources}
\label{sec:min}
Why do nervous systems in animals produce representations?  An answer 
is that by holding a representation an animal does not have to do the same
processing of information again and again.   Indeed repeated information 
processing might be the method by which many representations are created;
for example Jeannerod (1994) \cite{bi:jeannerod}
repeated actions seem to improve motor control,
in the words of Decety et al (1994) \cite{bi:decety}
\begin{quote}
Mental practice involves rehearsal of 
neural pathways related to cognitive stages of motor control.
\end{quote}   
Other examples which suggest that the representations 
used in imagination are the 
same as those received externally are:  i) the dual visual representations 
discussed by Farah (1988) \cite{bi:farah},  
see the previous section \ref{sec:stereo},  
and ii) the combined use of visual and 
auditory imagery when reading to oneself.   The principle of minimum 
duplication stands in stark contrast to the ideas of modularity where for 
example there is no visual and auditory interaction until the information 
reaches the ``cognitive center''.   Strict modularity is in any case unlikely 
for many reasons, three of which are:  i) the interaction of seeing and 
hearing words McGurk and McDonald (1976) \cite{bi:McGMcD},  
ii) the orienting motor response 
which occurs at the third relay in human auditory processing 
Garman (1990) \cite{bi:garman} p.62,  
iii) Frog "bug-perceivers" Garman (1990) \cite{bi:garman} p.69,  
appear already 
to occur in the retina before any subsequent visual information processing 
occurs.   The principle of minimum duplication is a minimality requirement 
in the same way as Ockam's razor (requiring the minimal explanation for 
scientific data) in the philosophy of science,  however ultimately it might 
be possible to formulate it as a minimality requirement like the principle 
of least action in physics.
\newpage
\section{Name Strategy in Colour Perception}\label{sec:colour}
\subsection{Berlin-Kay Colour Ordering}\label{sec:berlin}
The perception of colour often involves the deployment of a colour name 
strategy.  The effect of this is to alter the way the colour is perceived.
The five principles of colour perception have been given by 
Brown (1976) \cite{bi:brown}:
\begin{quote}
{\bf 1}. The communicability of a referent in an array and for a particular
community is very closely related to the memorability of that referent
in the same array and for members of the same community.
\newline
{\bf 2}. In the total domain of colour there are eleven small 
focal areas in which
are found the best instances of the colour categories named in any particular
language.   The focal areas are human universals,  but languages differ in the
number of basic colour terms that they have;  they vary from two to eleven.
\newline
{\bf 3}. Colour terms appear to evolve in a language 
according to the Berlin-Kay
(1969) \cite{bi:BK} universal partial ordering:
\end{quote}
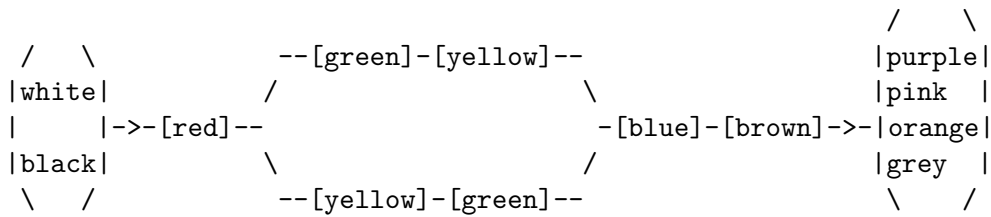
\begin{verbatim}                               
                                                          /    \
 /   \            --[green]-[yellow]--                   |purple|
|white|          /                    \                  |pink  |
|     |->-[red]--                      -[blue]-[brown]->-|orange|
|black|          \                    /                  |grey  |
 \   /            --[yellow]-[green]--                    \    /

\end{verbatim}
\begin{quote}
{\bf 4}.Focal colours are more memorable,  easier to recognize,  
than any other colours,  
whether the subjects speak a language having a name for the colour or not.
\newline
{\bf 5}.The structure of the colour space determined by 
multi-dimensional scaling 
of perceptual data is probably the same for all human communities and it is 
unrelated to the space yielded by naming data.
\end{quote}
\newpage
\subsection{The Sapir-Worf Hypothesis}
The five principles of colour perception can be used to test formulations
of the Sapir-Worf hypothesis.   This hypothesis is sometimes formulated Brown 
 (1976) \cite{bi:brown}, and Kay and Kempton (1984) \cite{bi:KK} 
as three separate hypotheses
\begin{quote}
{\bf S-W.I}   Structural differences between languages will,  in general,  be 
paralleled by non-linguistic cognitive differences,  of an unspecified sort,  
in the native speakers of the two languages.
\newline
{\bf S-W.II}   The structure of anyones native language strongly 
influences the 
world view he will acquire as he learns the language.
\newline
{\bf S-W.III}  The semantic system of different languages vary 
without constraint.
\end{quote}
\subsection{Name Strategy}
Kay and Kempton (1984) \cite{bi:KK} p.75 
define colour name strategy as follows:
\begin{quote}
According to the [colour] name strategy hypothesis,  the speaker who is 
confronted with a difficult task of classificatory judgement may use the 
LEXICAL classification of the judged objects as if it were correlated with 
the required dimension of judgement even when it is not,   so long as the 
structure of the task does not block the possibility.
\end{quote}
What this 
essentially means is that for colour judgement tasks the word for the 
colour may be deployed rather than the colour itself,  provided that the 
nature of the task allows for this.   This use of the word for a colour 
rather than the colour itself is a name strategy.   The significance of 
name strategy from a cultural point of view is that it allows {\bf S-W.I} 
to be tested.   
This can be done by using a language community in which all eleven 
focal colours have not been identified so that the language users do not have 
the option of deploying a name strategy.   North American native languages 
have one word for blue and green,  this is referred to as ``grue'' 
- thus these speakers do not have the possibility of deploying 
the words ``blue'' and ``green''
in judgement tasks so that they can be used as controls.   The scale of colour
differences between focal blue and focal green,  for the purposes of 
perception is measured not by wavelength,  but by a previously ascertained 
difference scale.   Kay and Kempton's experiments show that when the 
experiment involves the words blue and green,  
English speakers consistently and 
measurably assume colours near the blue/green boundary to be further toward 
the colour foci than they are;  however when colour words do not occur 
this does not happen.   They conclude that a naming strategy is involved 
which induces perceptual colour to move toward the colour foci.  
\subsection{Remarks On Colour Perception}
The existence of name strategy as discussed above can be interpreted 
as showing that thought can exist without words,  but when words do occur 
they can influence thoughts in a quantifiable way.   The effect of moving 
away from a difference boundary occurs not only in anthropology and 
psychology but in life science also:  Lack (1947) \cite{bi:lack} 
points out that two heavy billed 
finches Geospiza fortis and Geospiza fulinginasa,  
both have bills of approximately 
equal size when they are the only heavy billed finch on an island,  however
when they both occur on the same island the bill of one is much larger than
the other.   Brown (1976) \cite{bi:brown} p.152 
after noting the existence of colour foci 
universals and the above five properties of colour perception states
\begin{quote}The
fascinating irony of this research is that it began in a spirit of strong
relativitism and linguistic determinism and has now come to a position of 
cultural universalism and linguistic insignificance.
\end{quote}
This suggests the
possibility that there are universal perceptions and that it is the words
superimposed upon them that enhance the cluster properties mentioned by
Wittgenstein (1958) \cite{bi:wittgenstein} p.31;   
such cluster properties are characteristic of
concept formation,  see for example Hampton (1979) \cite{bi:hampton}  
and Nuyts and Pederson (1997) \cite{bi:NP},
and occur not only for
humans but also for other animals,  for example pigeons have a concept of 
what a tree is,   see for example Hernstein (1985) \cite{bi:hernstein}.
Note that,  Scriven (1997) \cite{bi:scriven},  Black,  Brown,  White,  Grey,
and Green are prevalent surnames,  but Blue,  Red,  Yellow,  
and Purple are not;
this seems to be also independent of the Berlin-Kay ordering.
\newpage
\section{Name Strategy in Object Perception}\label{sec:objpc}
\subsection{Object Perception}
It has been noted by Carmichael,  Hogan, and Walter (1932) \cite{bi:CHW},  
that language has an effect on visually perceived forms.  They constructed a 
number of experiments involving drawn figures that are midway between two 
clearly delineated objects,  some of their examples include:  a figure which 
is half way between a crescent moon and the letter ``c'',  and a number which 
is half way between a two and an eight, etc \ldots.   They claim on p.83,  
that their experiments:
\begin{quote}tend to substantiate the view that not the visual form 
alone,  but the method of its apprehension by the subject determines,  at 
least in certain cases,  the nature of its reproduction.
\end{quote}
and their final conclusion on p.83,  is that: 
\begin{quote}to some extent at least,  the reproduction of 
forms may be determined by the nature of the words presented orally to the 
subject at the time that they are first perceiving specific visual forms.
\end{quote}  
An example of a similar effect is the ink-blot technique of making letters 
ambiguous,  Lindsay and Norman (1972) \cite{bi:LN},  
this involves part of a letter in a 
display of a word being obliterated by an ink-blot;  the presence of the rest 
of the word disambiguates (i.e. removes the ambiguity in the choice of 
letter) the letter so that it is perceived as a letter which produces 
a word rather than a non-word.  In other words object name strategy 
is often deployed in the apprehension of visual objects.   In the 
view adopted here these are two examples of the similar phenomena to that 
found in colour perception;  to wit when language is present it can adjust 
thought via a naming strategy.   The difference is that these two object 
perception examples are not switched;  and as they can be subject to subsequent
adjustments they are not encapsulated either.
\newpage
\section{Memory Chunking}\label{sec:memch}
\subsection{Recalling Re-coded Events}
Short term memory has a capacity to re-call about seven distinct elements
of information,  Miller (1956) \cite{bi:miller}.  
Miller cites much evidence for this:  from sound perception,  
where untrained people have the ability to recognize 
about  six different pitches and five degrees of loudness;  musically trained 
people can recognize a greater number of distinct pitches.  There are similar 
results for the ability to recognize other uni-dimensional stimuli,  which 
means a single stimuli with out interference from other effects.   For 
multi-dimensional stimuli,  where there is more than one spectrum of features 
which distinguishes a stimuli,  there is a relationship between the number of 
stimuli present and the number of objects that can be identified;  inspection 
of Miller (1956) \cite{bi:miller}
Fig.6 suggests that the number of objects that can be 
identified is proportional to the hyperbolic tangent of the number of stimuli 
present.   Miller suggests that the mechanism by which this is achieved is by 
re-coding the sequence of information into chunks,  and this is referred to as
a chunking strategy;  he says,  Miller (1956) \cite{bi:miller} p.93:
\begin{quote}There are many ways to do
this re-coding, but probably the simplest is to group the input events,  apply
a new name to the group,  and then remember the new name rather than the 
original events.
\end{quote}   This re-coding strategy is chunking strategy in memory, 
which is here called memory chunking.   Memory chunking is not encapsulated as
it is possible for memories to suffer from subsequent interference.   Memory
chuncking is not switched.
\newpage
\subsection{Memory for Chinese Words and Idioms}
This mechanism seems to be related to the naming strategy of the previous 
sections,  the connection between the two effects is illustrated by the study 
of short term memory capacity for Chinese words done by 
Zhang and Simon (1985) \cite{bi:ZS}.
Chinese written characters exist in two varieties,  one type of characters
has a name the other does not.   Zhang and Simon asked some Chinese speakers
to remember eight sets of the two sorts of characters.   The subjects could 
immediately recall about six characters with names,  but short term memory
capacity for Chinese radicals not possessing common pronounceable names is 
about two or three items.  Thus short term memory appears,  on occasion, to 
adopt a name strategy similar to the name strategy of the two proceeding
sections;   however the new name referred to by Miller might not always 
be a specific word,  it might be some other label.   To see this consider 
short term memory in musical performance,  where what is remembered may 
be specific notes,  or a specific scale etc.,  chunking strategy is 
still involved but it is not now a name strategy.  Therefore the term 
stereotyping is used to cover all three phenomena.  
\newpage
\section{The Implications for Traffic Accidents}\label{sec:traffic}
\subsection{Colour Perception and Traffic Accidents}
\begin{verbatim}

    Table:  Rate of Involvement in Traffic Accidents of Cars by Colour.

            Number of Accidents per 10,000 Cars.

            Adapted from Table 7C.HMSO91.

Black White Red Blue Grey Gold Silver Other Beige Green Yellow Brown
176   160   157 149  147  145  142    139   137   134   133    133

Two-colour cars are classified by their main colour.
Other = Bronze,Pink,Orange,Purple,
        Maroon,Turquoise,Multi-coloured,&Unknown.
\end{verbatim}
The above table shows the rate,  per 10,000 licensed cars,  of accidents
classified by colour.   There is a striking resemblance of the three most
accident prone colours namely black,  white,  and red to the Berlin-Kay 
diagram;  and some resemblance among other colours,  however accidents of 
blue cars seem to be anomalously high compared to green,  yellow,  and 
brown cars.   Munster and Strait (1992) \cite{bi:MS} note:
\begin{quote}The data probably say more 
about differences in the types of car and driver represented in the colour 
groups than differences in the inherent safety of the car colours.  It is 
possible that certain colours are more popular among groups of drivers with 
a higher risk of accident such as young drivers or company car users.  Some 
colours may tend to be associated with particular makes and models of car.
No doubt factors such as visibility also affect accident risk,  but it would 
be difficult to distinguish their effects from those of the driver and 
vehicle,  using the available data.
\end{quote}
From the point of view presented 
in this paper,  colours which have a greater propensity to be involved in 
a colour name strategy are more likely to coincide with the colour of a 
car involved in an accident.   To put this another way:  people who drive a 
car of a well-defined colour,  are more likely to be involved in an accident.
There have also been studies into the connection between personality type 
and car colour, Marston (1997) \cite{bi:marston}.
\newpage
\section{Implications for Psycholinguistics}
\label{sec:psy}
\subsection{The Interaction Model}
\label{sec:int}
There is evidence that the reception of speech is interactive,  for
example there is evidence that seeing the speaker speak influences the word 
heard,  McGurk and McDonald (1976) \cite{bi:McGMcD};  
also there is evidence that verbal
speech production interacts with gesture,  McNeil (1985) \cite{bi:mcneil},  
and various 
other physiological activities Jacobson (1932) \cite{bi:jacobson} p.692,  
and these observations 
suggests that the verbal part of speech production is interactive.  A model 
which allows both phonological and semantic influences to interact is the 
psycholinguistic interaction speech production model,  
c.f. Stemberger (1985) \cite{bi:stemberger}.   
In this model,  when the word feather is activated a lot of other 
words are also activated with varying weights according to how closely they 
resemble feather.   This can be described by the diagram   
Stemberger (1985) \cite{bi:stemberger} p.148 and text:
\begin{quote}
Semantic and phonological effects on lexical access.
...an arrow denotes an activating link,  while a filled circle is an inhibitory
link.   A double line represents a large amount of activation,  a single solid
line somewhat less,  and a broken line even less.   Some of the inhibitory 
links have been left out... for clarity.   The exact nature of semantic 
representation is irrelevant here,  beyond the assumption that it is 
composed of features;   ... a word in quotation marks represents its meaning.
\end{quote}
There is suppression (also called inhibition) across a level,  and 
activation up or down to the next level.   This model accounts for syntax 
by giving different weights to the different words so that words on the left 
come first.   Speech errors come from the noise in the system.   There are 
three kinds of noise.   The first is that the resting level of a unit node 
is subject to random fluctuations;  with the result that it is not the case 
that the unit nodes degree of activation remains at the base line level.
A fluctuation could produce a random production of a part of a word.   The 
second is that words that are used with a high frequency have a higher resting 
level,  and therefore reach activation threshold,  or "pop out",  quicker.   
This implies that there should be less error for these high frequency words;  
furthermore it implies that when real words occur as an error,  higher 
frequency words should occur as errors more often,  and this does not happen.  
The third is the so-called systematic spread of activation;   this means that 
the weights in the interaction allow an inappropriate activation of word.   
\subsection{Serial and Connectionist Models}
Serial models consist of a series of boxes arranged in strict order.   
These process a message into language and then articulate it.   The evidence 
cited in section \ref{sec:min} 
and at the beginning of section \ref{sec:int} shows that serial language 
perception and production is unrealistic.   Serial models begin with a 
"message level",  Garret (1980) \cite{bi:garret}, 
and thus at the beginning requires thought, 
in the form of semantic content,  to exist without talk,  in the form of 
tangible language.   In connectionist models of language and word production,
for example Seidenberg and McClelland (1989) \cite{bi:SM} p.527,  
the semantic aspects of the model are put in by hand.
\newpage
\subsection{Language Production}
The very words "model of language production"  assume that some starting 
structure is metamorphosed into some finishing structure.   In the present
case it is clear that the finishing structure is speech and other forms of 
language,  but it is not clear what the starting structure is.   In other 
words,  what language is supposed to be produced from.   Now we consider how 
this appears to be done for serial,  connectionist,  and interactive models 
respectively.   In serial models,  Garret (1980) \cite{bi:garret},  
the starting structure 
is the message level;  here there are collections of amorphous thoughts 
which the production process crystalizes into language.  Now this begs an 
assumption,  because it implies that thought can exist prior to and/or 
independently of words.   There appears to be a lot of scope as to 
how meanings of words can be accommodated within connectionistic models,  for 
example Rumelhart and McClelland (1986) \cite{bi:RM} p.99 say:
\begin{quote}
there is an implicit 
assumption that word meanings can be represented as a set of sememes.
This is a contentious issue.   There appears to be a gulf between the 
componential view in which meaning is a set of features and the structuralist 
view in which the meaning of a word can only be defined in terms of its 
relationships to other meanings.   Later ... we consider one way of 
integrating these two views by allowing articulate representations 
to be built out of a number of different sets of active features.
\end{quote}    
The componential view is a sort of {\it correspondence theory} of meaning,
and similarly the structuralist view a {\it coherence theory} of meaning,
for definitions of these terms see Roberts (1998) \cite{bi:mdr98}.   
In interaction models the situation is a more complex than in serial models;  
here the starting structure is a "word in quotation marks which represents its 
meaning"  Stemberger (1985) \cite{bi:stemberger} p.148.   
This cannot really be a starting point 
because it requires that a word has meaning in isolation,  i.e. that context 
is irrelevant.   It has been argued by Frege,  Davidson,  and others that 
not only the material context but also the context of a word in a 
sentence is necessary for an unambiguous assignment of meaning to a word.   
An example of an ambiguous word is ``bank'',  what sort of bank is being
referred to can be disambiguated by a sentence or perhaps some other source 
of information.   An example of context ambiguity is:  ``1st speaker to 2nd:
Why don't you become a philosopher?''  2nd speaker: ``Don't have anything to do
with philosphers,  they are underhand.''   3rd speaker:  ``Why don't you become
a philosopher.''   Sentence themselves are often ambiguous because they contain
ellipses and indexicals,  again they have to be dis-ambiguated by being
put in the perspective of a wider context.   This gives rise to a ``holistic'' 
view of the meaning of a word,  this has been expressed by 
Davidson (1984) \cite{bi:davidson} p.22 as follows:
\begin{quote}
If sentences depend for their meaning on their structure,
and we understand the meaning of each item in the structure only as an 
abstraction from the totality of sentences in which it features,  then 
Frege says that only in the context of a sentence does a word have meaning;  
in the same vein he might have added that only in the context of the language 
does a sentence (and therefore a word) have meaning.
\end{quote}   
Again it would appear that the starting structure is a thought;  
thus the models all assume 
that a thought occurs first and is processed into words, i.e. that thought 
and words are not identical,  but are independent entities.   
Davidson (1984) \cite{bi:davidson}
addresses this question,  however he seems just to be concerned that if 
thought can occur without words,  then this is also subject to radical 
interpretation,  for example on p.157:
\begin{quote}
the chief thesis of this paper 
is that a creature cannot have thoughts unless it is an interpreter of 
the speech of another.
\end{quote}
this suggests to me that Davidson thinks that
words and thoughts are concomitant simultaneously.
\subsection{Sememes and Phonemes verses Thought and Talk}
The preceeding suggests that generally psychologists require sememes/thought
to exist independently and prior to phonemes/talk;  whereas philosophers are 
wary of thoughts existing independently.   This is summed up by Wittgenstein's
remark on where one cannot speak one cannot think.   This is perhaps because 
of their different attitudes to studying communication.   Psychologists are 
mainly concerned with explaining some particular aspect of language.   
Philosophers are most interested in accuracy and rigour of thought,  
if these are not written down they are hard to refute.
\newpage
\subsection{Justification of Prior Thought}
The assumption of the models,  namely that thought can exist without 
words,  is justified for at least five reasons.   
The {\it first} is because of 
the existence of the tip of the tongue (TOT) phenomena,  
Brown and McNeill (1966) \cite{bi:BM}.  
This occurs when people appear to be at a loss for the word that 
they are looking for.   The {\it second} is because of the nature of musical 
instrument tuition,  where it is possible to think about phrasing,  etc \ldots,
without recourse to words;  this is contrary to Davidson's assertion,
that a creature cannot have thoughts unless it is an interpreter,  quoted 
at the end of section \ref{sec:quan},  
irrespective of whether musical conventions and 
performance are subject to radical interpretation,  simply because notes 
are not speech.   The {\it third} is that animals have brains 
and so are able,  at least to some extent,  to have something similar to 
thoughts,  see for example Terrace (1985) \cite{bi:terrace} 
and Premack (1988) \cite{bi:premack}.   
The {\it fourth} is semantic amnesia studies,  
DeRenzi et al (1987) \cite{bi:derenzi};  their patient L.P.
appears to have selective damage which suggests that semantics and phonetic
are separable,  for example on p.579:
\begin{quote}
The linguistic performance of L.P.
was marked by a striking dissociation between impairment of semantic
knowledge and the preservation of phonetic,  grammatic,  and syntactic rules.
\end{quote}
The {\it fifth} and clearest reason is because of the 
existence of name strategy,  
this not only shows that thoughts can exist independently of words,  but 
futhermore colour name strategy allows quantitative statements of the degree 
of independence to be made;  in colour name strategy there is a perceived 
movement away from the boundary between colours,  and the amount of this 
movement could provide such a measure.   There is the question of the
temporal order of the independence,  thought before talk seems to be the 
assumption of speech production models and seems to occur,  but there is the 
possibility of the reverse which can be facetiously put as talking without 
thinking.   In colour perception the existence of a switched focal colour 
word in a speakers repertoire adjusts the perception of colour,  so that it 
appears that talk,  in the form of name strategy,  can occur prior to thought.
\newpage
\section{Implications of for the Philosophy of Language}
\label{sec:radical}
\subsection{Radical Interpretation}
\label{sec:rad}
Philosophy of language is concerned with how language has meaning.   
A current view,  Davidson (1984) \cite{bi:davidson} 
is that meaning is obtained through 
radical interpretation of sentences;  
or perhaps,  Lewis (1974) \cite{bi:lewis} a larger 
structure.  Radical interpretation optimizes the correspondence between 
language and the world and also the self-consistency (or coherence) of 
the language.   I have given an outline of some aspects of my approach 
to it in Roberts (1998) \cite{bi:mdr98}
and this section follows on from there.   
In section \ref{sec:colour} it was shown that the existence of a colour name 
strategy reduces the correspondence between the true colour of 
a chip and its perceived colour;  thus language does not help 
the mind in producing an accurate representation corresponding 
to the world,  in fact in hinders it.   In other words the use of 
a more polished language is obscuring truth.   The assignment of truth
values to statements is essential in the traditional picture of radical
interpretation.   Thus the question arises:  is the existence of name 
strategy compatible with a modified form of radical interpretation,  
and if so exactly how is it necessary to change radical interpretation?
\subsection{Accuracy of Correspondence}
A specific and neat illustration of the dilemma is posed by the question:
Why have two words ``Blue'' and ``Green'' 
when all they achieve is a less accurate
representation of the colours present?   This is part of a question that can 
be successively generalized to include other colours;  also it is related to 
why have memory chunking when this might cause a slight loss of accuracy,  
and furthermore as to why have stereotyping in general.   A point to note is 
that usually these classifications are unconscious,  and that this allows them
to occur at great speed -  thus aiding swift communication.
\newpage
\subsection{Quantity of Transferred Information when in Re-coded Form}
\label{sec:quan}
The answer to the dilemma is that it allows a large amount of information
to be communicated,  even though there is a small cost in accuracy.   It might
be more accurate to not have separate words for blue and green,  and then 
to indulge in lengthy comparisons every time a distinction between blue and 
green objects is required;  but if large amounts of information is to swiftly 
communicated it is beneficial to have two separate words.   For example,  it 
is advantageous for a language community to be able to say:  "Blue Gavagai 
are tasty, but Green Gavagai are poisonous"  rather than:  "Gavagai that are
grue like the sky are tasty,  but Gavagia that are grue like the grass are 
poisonous.".   More formally,  for a specific case of colour perception
\ref{sec:berlin},  
the first principle,  which states that the communicability of a referent 
is related to its memorability,  illustrates the mechanism by which larger 
quantities of information can be more quickly communicated.   A name strategy 
is chosen so that a large amount of information about the colour of an object
can be easily remembered;  and also be quickly and memorably communicated,  
despite the fact that it incurs a loss in the accuracy and truth of statements.
Thus economy of expression accounts for the use of one word instead of two.
There is still the problem of why there are the eleven switched colour 
perception centers.   Why not have ten or fifteen switches colours,  or none
at all?   These are human universals as given by the second principle of 
colour perception.   A possible explanation is that the switched colours 
provide a framework within which a huge number of colours can be described.
No switches would not give this framework.  The reason that there are eleven
of them might be that this is the maximum number that the brain can implement.
\newpage
\subsection{Circuitous Correspondence}
The foregoing presents a problem for the philosophy of language;  
because it shows that words are only in indirect correspondence with 
the world.   This problem is approached by attempting to explain indirect 
correspondence by using standard techniques and seeing if these lead to a 
plausible explanation;  the attempt consist of requiring that indirect 
correspondence is really a {\it circuitous correspondence}.   
By this is meant a pair $\vb+{+D,P\vb+}+$.   
$D$ is a direct correspondence and $P$ a procedure with which the 
indirect correspondence $I$ can be recovered.   Notationally this could be 
expressed as $D \vb+&+P\rightarrow I$.   
The usual view of radical interpretation requires that 
sentences are bivalent,  or two-valued;   this means that the truth values 
"True" or the truth value "False" can be assigned to them. 
Here the possibility of assigning any quantity to (or labeling, or marking,
or tagging \ldots) sentences is referred
to them as being {\it flagged}.   
The existence of a name strategy,  especially in 
colour perception,  shows that there is no unique way of achieving this;  for 
example,  quantitative experiments could be constructed in which subjects 
would be required to state how blue or green a chip is,  the quantitative 
results would depend on whether a name strategy was being deployed,  then true
or false sentences could be constructed to describe these quantitative 
differences,  
but these would have different truth values depending on whether a 
name strategy was being deployed.  This ambiguity of truth values necessitates
that a disambiguating procedure (i.e. a procedure which removes any amgibuity)
be given.   The form that such a procedure could take is the re-writing,  
hopefully in a bounded finite number of steps $B$,  of the original sentences 
which are bivalent.   An example of a re-writing procedure is what is here 
called a {\it brittle} re-write procedure.   
This can be described as follows:  
suppose that sentences can be given a single definite value,  
c.f. Urquhart (1986) \cite{bi:urquhart},  
of say:
\be
U=\vb+{+false, almost false, almost true,  true\vb+}+;  
\ee
then the original 
set of sentences can be replaced by a larger set of sentences having just 
the values true or false,  this can be done by posing the four additional 
sentences:   "The previous sentence is  {False,  Almost False,  Almost True,  
True}"  all four of which are true or false.   Another example,  is to suppose 
that,  instead of the sentences being bivalent it is possible to assign a set 
of labels or quantities $Q$ to a sentence,  for instance 
\be
Q = \vb+{+Happy~ and~ Good,  Happy~ and~ Bad,  
Sad~ and~ Good,  Sad~ and~ Bad \vb+}+;
\ee  
again for additional sentences,  
such as:  "The previous sentence has the quality $Q$",  
can be constructed which 
are bivalent.   This approach really falls under the scope of Davidson's 
(1984) \cite{bi:davidson} p.133  program,  
it is just the requirement that there should be a 
procedure for matching sentences without logical form (i.e. bivalent sentence)
the gerrymandered part of a language where sentences do have logical form.  
It might be thought straightforward,  for the circuitous correspondence of 
colours,  to implement  such a procedure;  however first two problems have to
be overcome.   
The {\it first} is that the re-write procedure would involve real 
variables,  the parameters $P$ to be adjusted,  such as wavelengths,  take 
values $V$ on the real line  $\Re$ ,  i.e.$V(P) \varepsilon  \Re$;  
however it seems reasonable 
to assume that sentences and truth values can only be stated and assigned 
sequentially,  so that the total number of truth values available $N(T)$ is an 
integer valued quantity,  i.e.$N(T)\varepsilon Z~ $;  thus there 
is the problem of how
to segment $V(P)$ so that $N(T)$ covers all cases,  this is an example of the 
{\it segmentation problem},  discussed in section \ref{sec:cont}.   
The {\it second} is that the re-write
procedure has to be constructed {\it post-hoc},  
after data about responses has been
given,  but extralinguistic assignment (or flagging) of truth values to 
sentences is immediate;  this is an example of the 
{\it extralinguistic assignment problem} 
discussed in section \ref{sec:extra}.   
These two problems suggest that the traditional view of 
radical interpretation needs radical revision.
\subsection{The Extralinguistic Assignment Problem}
\label{sec:extra}
In order for radical interpretation to work various principles,  such as
the {\it principle of charity},  have to be assumed.   
I have previously,  Roberts (1998) \cite{bi:mdr98} 
explained my version of this which I called 
{\it cooperation dominance}.   
These principles contain the idea that people communicate truthfully 
more often than not,  and that details of their behaviour give indications 
of the truthfulness of their communication;  hence it is possible to use 
{\it extralinguistic information} to assign the truth values to sentences.   
These principles can be modified so that real valued measures of truth 
are assigned, instead of just the bivalent assignment of 
\be
B=\vb+{+true,false\vb+}+.
\ee  
However a problem arises because the existence of a name strategy shows that,
at the time an apparently true statement is uttered the quantity of truth is
not optimal;  that the statement is apparently true is {\it post hoc} 
as it assumes
that the listener posses the information,  consciously or unconsciously,  to 
transcribe it to the relevant true statement.   A way that this could be 
accommodated in the usual theory is be requiring that the transcription 
process is contained in the {\it passing theory} of the language;  but then 
it seems hard to justify referring to the extralinguistic information 
assigned as being "truth".    A better description is to assume that the 
extralinguistic assignment is of relevant information,  
c.f. Sperber and Wilson (1987) \cite{bi:SW};  
the traditional case is a sub-case of this,  
as most relevant information is true,  but also it is possible 
to state irrelevant true statements.   That truth is not the only quality
needed for facility of communication has been noted by 
Church (1956) \cite{bi:church} p.2-3 
\begin{quote}
For purposes of logic to employ a specially devised language,  a formalized
language as we shall call it,  which shall reverse the tendency of the natural
languages and shall follow or reproduce the logical form [here having the 
restricted meaning that statements can be assigned bivalent truth values
rather than be flagged by a formal system]  -  at the expense, 
where necessary,  of brevity and facility of communication.
\end{quote}
It is, see the P-model Roberts (1998b) \cite{bi:pmodel},  
possible to quantify information,  but quantifying relevant information 
presents problems,   perhaps the vehemence or Fregian force with which 
a sentence is stated could lead to such quantification.   Flagging by 
Fregian force is unrealistic as it implies that by shouting louder makes a
statement more true,  a better flag is relevant information.   Relevant 
information is discussed in the commentary on 
Sperber and Wilson (1987) \cite{bi:SW}.
The above relates meaning to relevant information,  supposing that 
relevant information is frequently used in communication relates 
meaning to use,  
compare Wittgenstein (1958) \cite{bi:wittgenstein} p.43
\begin{quote}
For the word 'meaning' it can be defined thus:  
the meaning of a word is its use in the language.
\end{quote}
\subsection{Flagging by Countenance}
What is the nature of the concomitant flagging which occurs when a 
statement is uttered?   The most important aspect of such flagging must 
be in the comparison with the passing theory of the language to preserve
coherence (interior flagging);  however at sometime exterior flagging must
occur.  Flagging occurs for purely auditory statements,  for example a tape
or telephone.   This flagging is typically of the mood,  age,  and sex of
the speaker,  and seems to give information in addition to statements rather
than be an aid to comprehension.   The main component of flagging for 
comprehension is probably visual.  
Boyle,  Anderson,  and Newlands (1994) \cite{bi:BAN} show that
having a visible speaker improves the efficiency of dialog and that this is 
more marked for young or inexperienced speakers.   This suggests that the 
extralinguistic information required is at least in part conveyed by the 
visual acts of the speaker;  as is it more marked for people learning the 
language more extralinguistic information is required here.   Clearly some
of visual information is in the form of demonstrable acts 
\be
D=\vb+{+ pointing~ out~ left/right,  ~up/down~ etc.\ldots\vb+}+,
\ee  
and gesture emphasis.  These could be used for flagging in the form 
\be
Q=\vb+{+degree~ of~ emphasis,  spatial~ and~ temporal~ location,\ldots\vb+}+,
\ee
but it is not clear where this list would end or the relative importance of
its components.  A neater way is to note that one of the main components
of visual communication are facial expressions,  and these can
be considered as contenders for the flags assigned.   Of course it is possible
to combine these with flags such as the above but with a loss of simplicity.
The advantage of using facial expressions can be taken to convey the most 
relevant information is that 
Etcoff and Magge (1992) \cite{bi:EM} find only five or six 
emotions that faces express:  happiness,  sadness,  fear,  anger,  disgust, 
and perhaps surprise.   Flagging using these five or six emotions produces an
assignment of extralinguistic information called flagging by $\Re^{6}$ facial
countenance.   Thus an alternative to sentences  being assigned bivalent 
truth values or measures of unspecific relevant information,  they can be 
assigned six real values 
\be
Q=\vb+{+happiness,  ~sadness,  ~fear,  ~anger,  ~disgust, ~surprise \vb+}+.
\ee   
These can be expressed in ascii smilies 
\ber
Q=\vb+{+&:-)&,  ~:-(,  ~:-F,~:-t,  ~:-*,  ~:-s \vb+}+\nonumber\\ 
= \vb+{+ &happy&~ smilie, ~sad~ smilie, ~bucked-tooth~ vampire~ with~
one~ tooth~ missing,\nonumber\\
cross ~&smilie&, ~ate ~something ~sour ~smilie, 
~smilie ~after ~a ~bizzare ~comment \vb+}+.
\ear  
Thus six real valued flags replace replace the one 
integer valued flag 
\be
Q=\vb+{+truth\vb+}+. 
\ee  
This assignation taken to be the driving 
force behind the optimization of correspondence between language and 
the world and hence radical interpretation.
\newpage
\section{Implications for the Philosophy of Mathematics}
\label{sec:maths}
\subsection{The Continuum Hypothesis and the Segmentation Problem}
\label{sec:cont}
In mathematics there is an assumption:  the {\bf Continuum Hypothesis} see for
example Maddy (1993) \cite{bi:maddy} and Hirsch (1995) \cite{bi:hirsch},  
that mathematical objects can have
continuous properties.  A major place where this hypothesis appears is that 
real numbers cannot be constructed from rational numbers necessitating a new 
assumption for real number construction.   An example of such an assumption
is given by Dedekind's axiom of completeness,  see for example 
Issacs (1968) \cite{bi:issacs}.
Hirsch (1995) \cite{bi:hirsch} p.146 notes as a possibility that
\begin{quote}
(b) Neurologists and 
psychologists learn enough about cell assemblies and cognition to make it 
scientifically certain that there could not possibly be any activity in the 
nervous system which would correspond to a truth value for the Continuum 
Hypothesis.
\end{quote}
What is about to be advocated here is something along these 
lines,  essentially it is argued that both truth and meaning are intrinsically
real valued.   
The {\it segmentation problem} is the problem of how to account for
the existence of and meaning attributed to the real numbers.   From the point
of view of traditional radical interpretation the number of times truth can be
assigned to natural language statements is integer valued,  and so does not 
allow for real numbers.   There are three parts to this problem:
\newline
i) how to 
account for the fact that real numbers as pure mathematical constructs have 
meaning, 
\newline
ii) how to account for the success,  meaning,  and use of real numbers
in the physical world,  
\newline
iii) how to account for the success,  meaning and use 
of real numbers in cognitive science.
\newline
In the previous paragraph a simple 
example,  which shows that the real numbers are necessary for assigning truth 
values,  is given by colour name strategy.   Similar problems arise with many 
other correspondences between languages and the world,  for example consider 
the statement: 
'$Jones~ is~ six~ feet~ tall$'.   
This is unverifiable,  the statement 
for which truth-values can be assigned is:  
'$Jones~ is~ six~  \pm \delta~ feet~ tall$'
where $\delta$ is some error,  usually of a statistical nature.  It is normally
taken that physical measurements are real valued quantities with real valued 
errors,  and therefore the real numbers are needed to describe the physical 
world.   It could be argued that a theory of everything ({\bf TOE}) 
Taubes (1995) \cite{bi:taubes},
would require that objects have only discrete properties,  
say by having length a multiple 
of the Planck length etc.,  but present quantum mechanics requires Hilbert 
spaces which in turn require properties of the real numbers,  so that a 
requisite {\bf TOE} would have to change this;  
in any case it seems unreasonable 
that philosophy of language and mathematics
should legislate the nature of fundamental 
physics.   A better reason for discarding the real numbers would
be to assume that humans have only a finite number of integer valued
brain states,  so that the assignment of a finite number of truth values 
should suffice for the assignment of meaning as perceived by the brain.
The indications are that real valued quantities are needed for describing
cognition and hence brain states.   Three examples of this are:
\newline
i) the accumulator theory of how people learn to count 
Wynn (1992) \cite{bi:wynn} p.323,  
this requires that 
people can perceive temporal duration,  a real quantity,  
and then when sufficient amount of this has accumulated 
set an integer valued quantity one higher,
\newline
ii) a clock measuring real valued temporal duration is needed for many skills 
involving timing Shaffer (1982) \cite{bi:shaffer},  
\newline
and iii)most perceptual models,  Massaro and Friedman (1990) \cite{bi:MF} 
involve real valued quantities.
\newline   
These three examples 
suggest that real valued quantities should be used for describing brain 
states;   however Hopfield (1984) \cite{bi:hopfield}
has shown that continuous neurons,  which 
is what real neurons are,  can often be described by two-state McCullock-Pitts 
neurons,  suggesting that for psychological measurements bivalent quantities 
might suffice.   The problem is in fact a worse than just requiring  
real numbers for physical and psychological measurements.    Real numbers 
occur in pure mathematics and a general account of meaning should account 
for meaning in such {\it a-priori} languages,   because apart from any other 
reason mathematics is good at describing the physical world,  Wigner (1960)
\cite{bi:wigner}.   
It appears that the {\it segmentation problem}
does not have a solution involving
only integers.   The pure mathematical reason for this is that accounting for 
the meaning of the real numbers cannot be done by assigning an integer number 
of truth values,  there is the requirement of a new axiom 
when constructing $\Re$.
To put this another way having discrete truth values is not sufficient,  even 
with {\it brittle} re-write procedures,  for $\Re$ to have meaning it must be 
possible to assign quantities $Q \varepsilon \Re$   to sentences.   
To put this another way 
sentences cannot be flagged by integer valued quantities,  they have to 
be flagged by real valued ones.   In traditional radical interpretation truth
is an {\it ab initio} concept,  
for maximum compatibility with this real valued
generalizations of truth are required.   Possible choices for real valued
truth are probabilities or fuzzy truth values.   
\subsection{Probabilistic and Fuzzy Tarski Truth Theory}
Radical interpretation is in part motivated by Tarski's truth theory.   
Usually this applies bivalent truth values to formal languages.   
Church (1956) \cite{bi:church} p.25 
and by implication Davidson (1984) \cite{bi:davidson} p.19  
seem to use the term 
``logical form''  where ``bivalence'' has been used here.   In other words
they require integer valued flags,   but here it is advocated that real valued
flags are necessary.   There are formalized languages which do 
not require that truth is bivalent but is real valued,  examples of 
these are fuzzy logic and probability theory.   As has been argued in 
the previous section real values should be used for describing truth 
in natural languages.  There appears to be no reason why Tarski's truth 
theory cannot be modified to include statements which involve probabilities 
or fuzzy truth values,  although this appears not to have been explicitly 
done in the literature.   Instead of defining a relation:
"Assignment a satisfies formula $F$ in structure $S$:
define a relation:
"Assignment a gives formula $F$ the value $p$ in the structure $S$"
in the probability case,  $p$ will be a real number between $0$ and $1$.   
\newpage
\section{Summary}
\label{sec:sum}
\subsection{Peroration}\label{sec:peroration}
So what does a speaker take a sentence such as "Snow is white." to mean?
Stripping away multitudinous provisos the essentially old view of radical
interpretation would be along the lines:  ``In a formal Tarski truth theory
the truth of such a statement would be ascertained by comparison with the 
statement in another formal language.   For learning a first natural 
language with no language to compare it to,  the truth of a previously 
unheard statement is judged by seeing how well it matches the listeners 
previous model of the language and the external information available.  
The external information available is taken more often than not to
be useful rather than misleading.   Having judged a statement to be true the
speaker then deduces that it is in accurate correspondence to the exterior
world.  Having information that accurately portrays the exterior world must 
entail knowing what it means;  meaning is implicated by the above reasoning
and is a derivative concept from it.''   So how does the picture of radical
interpretation presented here differ?  It differs in that truth is not a 
quantity that can be flagged to sentences.  The two reasons are:  i) bivalent
truth values could never account for the meaning of the real numbers $\Re$ 
and the continuous properties that they describe throughout human experience,  
ii) there appears to be little psychological evidence that 
learners of a language
flag sentences with truth values other real valued quantities seem more 
likely.   There are three main consequences of this:
\newline
1) not only is meaning a derivative concept,  truth is as well,  
\newline
2) real numbers have to be introduced 
{\it ab initio} rather than be constructed from the integers,  
integers are thought of as being a derivative concept,  
\newline
3) perhaps in principle mathematical or 
computational models,  or psychological data could test aspects of the above;
hopefully it moves the theory one step closer to being testable.
\newpage
\subsection{Conclusion}
\label{sec:conclusion}
In sections \ref{sec:colour},  \ref{sec:objpc},  and \ref{sec:memch}
three distinct phenomena were discussed:
in section \ref{sec:colour} the phenomena was colour name strategy,  
this occurs only for the eleven switched focal colours;   
in section \ref{sec:objpc} the phenomena 
was object name strategy,  this is not an identical phenomena to colour name 
strategy because there is nothing which directly corresponds to the fixed 
number of switched focal colours;   in section \ref{sec:memch} 
it was mentioned that memory 
chunking is aided by the use of a name;  these strategies are clearly related
and it was argued that the relationship is that of the diagram 
\ref{sec:dia} in the Introduction.   
It was argued in \ref{sec:moti} and \ref{sec:stereo} that all of these should 
be considered representations,  and that representations should be
hierarchically classified according to their properties.   
In sections \ref{sec:radical} and \ref{sec:maths}
some of the consequences of name strategy for problems in the philosophy 
of language and mathematics were outlined.   These were summarized in 
subsection \ref{sec:peroration}.
The technical results follow below:  It was noted that having two colour words 
blue and green aids swift communication although there can be a loss in 
accuracy.   The suggested solution to the segmentation problem was to require 
that real valued quantities (as opposed to the traditional view where a single
bivalent quantity - truth) should be assigned to sentences 
and used throughout in accounts of meaning.   
Thus the {\bf Continuum Hypothesis} in mathematics 
is evaded as it is only possible to have real valued truth.   It was argued 
that the extralinguistic assigned qualities,  used in accounts of meaning,  
should not be unconditional truth,  
but rather be flagged by information relevant to comprehension,  
perhaps as signified by the six facial emotions;  
this is here referred to as flagging by countenance.   The existence of name 
strategy provides the clearest,  of several,  indication that thought and 
talk are separate entities;  this partially vindicates a hidden assumption 
in speech production models \ref{sec:psy},  
where it is not unambiguously stated what the 
starting structure which speech is supposed to be produced from is.
\section{Acknowledgements}
I would like to thank Mr.R.E.Strait for forwarding information on
the colour of cars involved in traffic accidents,   and also Prof.W.
Hodges for communication about Tarski's truth theory.
This work has been supported in part by the South African
Foundation for Research and Development (FDR).

\end{document}